\PassOptionsToPackage{table}{xcolor}
\documentclass{article} 
\usepackage{iclr2024_conference,times}

\usepackage{amsmath,amsfonts,bm}

\def\eqref#1{equation~\ref{#1}}

\def\1{\bm{1}}

\DeclareMathAlphabet{\mathsfit}{\encodingdefault}{\sfdefault}{m}{sl}
\SetMathAlphabet{\mathsfit}{bold}{\encodingdefault}{\sfdefault}{bx}{n}

\usepackage{hyperref}
\usepackage{url}
\usepackage{graphicx}

\usepackage{booktabs}
\usepackage{tablefootnote}
\usepackage[flushleft]{threeparttable}

\usepackage{lipsum}
\usepackage{bbding}

\usepackage{color}
\usepackage{enumitem}

\usepackage{multirow}
\usepackage{ulem}

\usepackage{tcolorbox}       
\usepackage{amsfonts}       
\usepackage{nicefrac}       
\usepackage{microtype}      

\usepackage{float}

\definecolor{Gray}{gray}{0.8}

\title{Step-DPO: Step-wise Preference Optimization for Long-chain Reasoning of LLMs}

\author{
Xin Lai$^{1}$
~
Zhuotao Tian$^{2}$
~
Yukang Chen$^{1}$
~
Senqiao Yang$^{2}$ 
~
Xiangru Peng$^{2}$
~
Jiaya Jia$^{1,3}$
\\[0.2cm]
$^1$The Chinese University of Hong Kong~~
$^2$Harbin Institute of Technology (Shenzhen)~~
$^3$SmartMore
}

\iclrfinalcopy 
\begin{document}

\maketitle

\begin{abstract}
Mathematical reasoning presents a significant challenge for Large Language Models (LLMs) due to the extensive and precise chain of reasoning required for accuracy. Ensuring the correctness of each reasoning step is critical. To address this, we aim to enhance the robustness and factuality of LLMs by learning from human feedback. However, Direct Preference Optimization (DPO) has shown limited benefits for long-chain mathematical reasoning, as models employing DPO struggle to identify detailed errors in incorrect answers. This limitation stems from a lack of fine-grained process supervision. We propose a simple, effective, and data-efficient method called Step-DPO, which treats individual reasoning steps as units for preference optimization rather than evaluating answers holistically. Additionally, we have developed a data construction pipeline for Step-DPO, enabling the creation of a high-quality dataset containing 10K step-wise preference pairs. We also observe that in DPO, self-generated data is more effective than data generated by humans or GPT-4, due to the latter's out-of-distribution nature. Our findings demonstrate that as few as 10K preference data pairs and fewer than 500 Step-DPO training steps can yield a nearly 3\% gain in accuracy on MATH for models with over 70B parameters. Notably, Step-DPO, when applied to Qwen2-72B-Instruct, achieves scores of 70.8\% and 94.0\% on the test sets of MATH and GSM8K, respectively, surpassing a series of closed-source models, including GPT-4-1106, Claude-3-Opus, and Gemini-1.5-Pro. Our code, data, and models are available at \url{https://github.com/dvlab-research/Step-DPO}.
\end{abstract}

\section{Introduction}

\vspace{-0.6cm}
\begin{figure}[H]
\begin{center}
\includegraphics[width=0.91\linewidth]{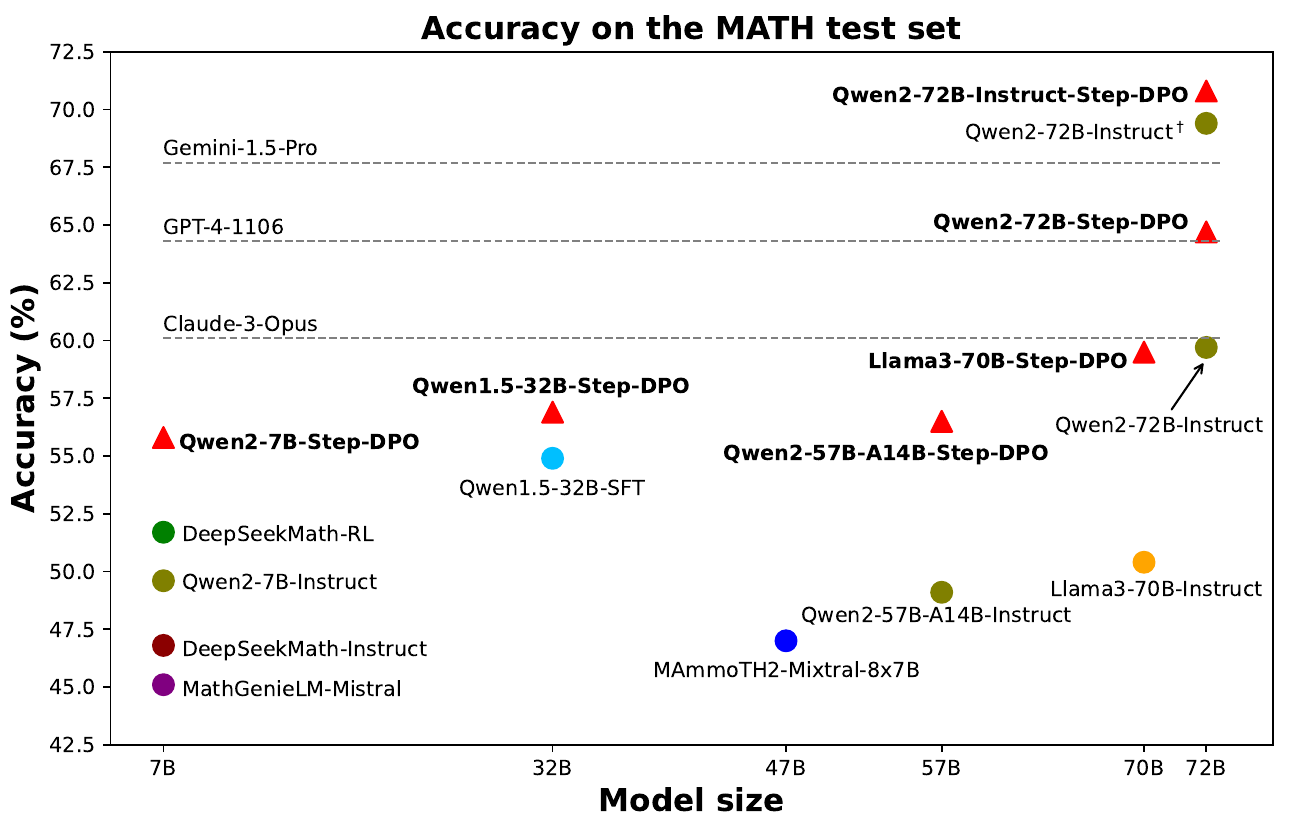}
\end{center}
\vspace{-0.6cm}
\caption{Accuracy on the MATH test set across models fine-tuned by Step-DPO and other state-of-the-art models. $^\dagger$: reproduced result using our prompt.}
\label{fig:acc_scatter}
\end{figure}

\begin{figure}[t]
\begin{center}
     \begin{minipage}  {0.49\linewidth}
        \centering
        \includegraphics [width=1\linewidth,height=1\linewidth]
        {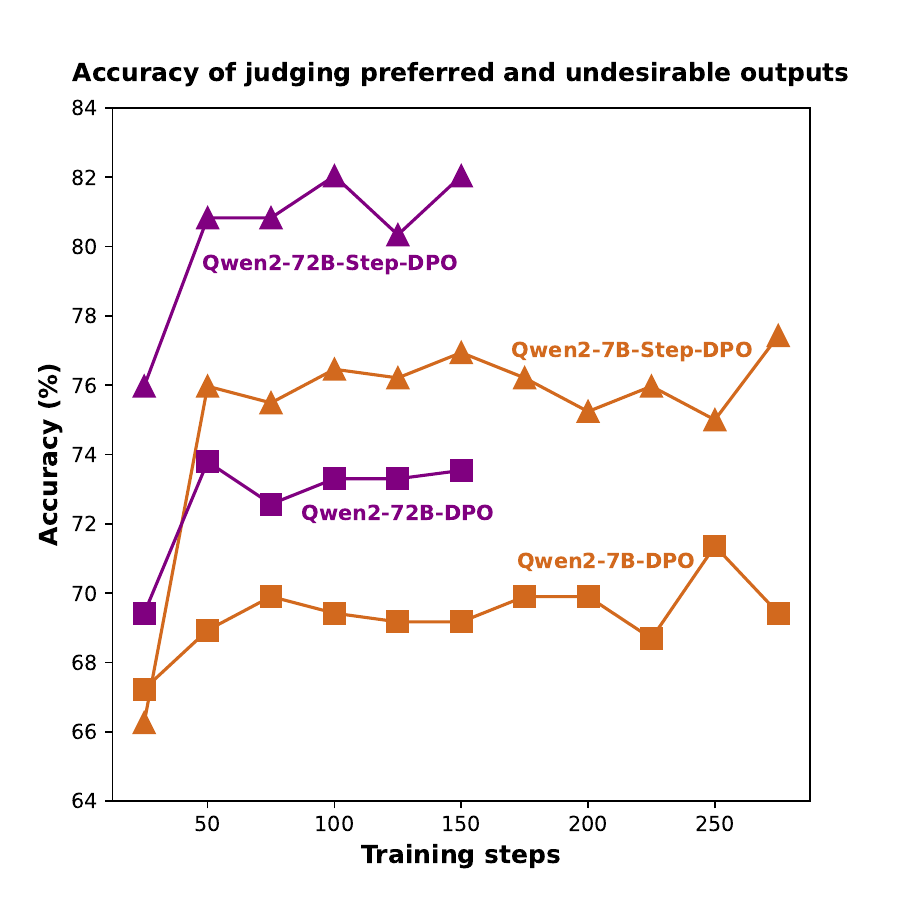}
    \end{minipage}
     \begin{minipage}  {0.49\linewidth}
        \centering
        \includegraphics [width=1\linewidth,height=1\linewidth]
        {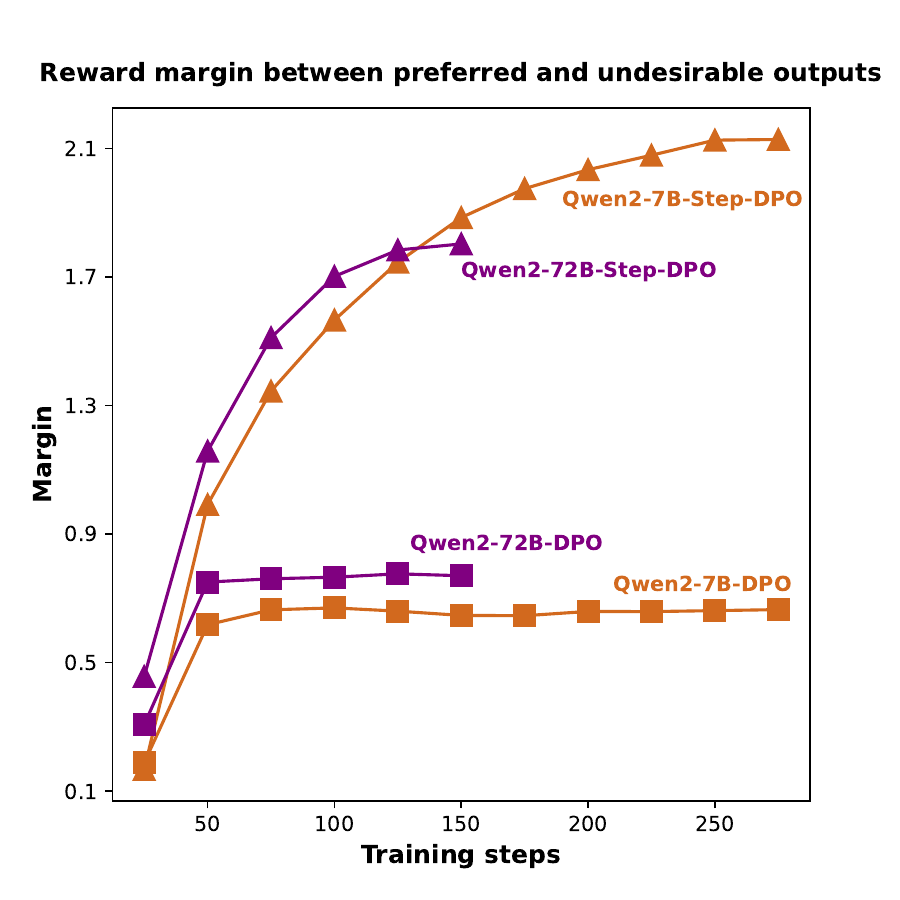}
    \end{minipage}
\end{center}
\vspace{-0.4cm}
\caption{\textbf{Left}: Accuracy of judging preferred or undesirable outputs on the validation set during training. \textbf{Right}: Reward margins between preferred and undesirable outputs on the validation set during training. More details about these experiments are given in the appendix.}
\label{fig:line}
\end{figure}

Mathematical reasoning is recognized as a critical long-chain reasoning ability in Large Language Models (LLMs). This task is particularly challenging due to the often extensive chain of thought required, which can include numerous reasoning steps. Any error in these steps can lead to an incorrect final answer.

Numerous studies~\citep{yu2023metamath,luo2023wizardmath,yue2023mammoth,liu2024augmenting,lu2024mathgenie,li2024common,shao2024deepseekmath,xin2024deepseek,yue2024mammoth2,tang2024mathscale} have proposed various data augmentation techniques during the supervised fine-tuning (SFT) stage to enhance alignment. However, models in the SFT process are prone to hallucinations, resulting in saturated performance. A potential reason for this, as highlighted in \cite{hong2024orpo}, is that as the probability of preferred outputs increases, so does the probability of undesirable ones. This phenomenon makes the model more likely to make errors in long-chain reasoning. Therefore, it is essential to develop methods to suppress the likelihood of undesirable outputs.

Recently, Direct Preference Optimization (DPO)~\citep{rafailov2024direct} has been proposed for alignment using pair-wise preference data and is popular due to its simplicity. Despite its effectiveness in chat benchmarks~\citep{tunstall2023zephyr,zheng2024judging}, DPO offers minimal benefits for long-chain mathematical reasoning. As shown in Fig.~\ref{fig:line} (left), models using vanilla DPO perform poorly in distinguishing between preferred and undesirable outputs, failing to identify errors in rejected answers. Additionally, Fig.~\ref{fig:line} (right) shows that the reward margin (i.e., the gap between the rewards of preferred and undesirable outputs) is limited for models using vanilla DPO and plateaus with further training. These findings indicate that models fine-tuned with vanilla DPO cannot pinpoint detailed errors in incorrect answers, hindering the improvement of reasoning abilities.

\begin{figure}[t]
\begin{center}
\includegraphics[width=1.0\linewidth]{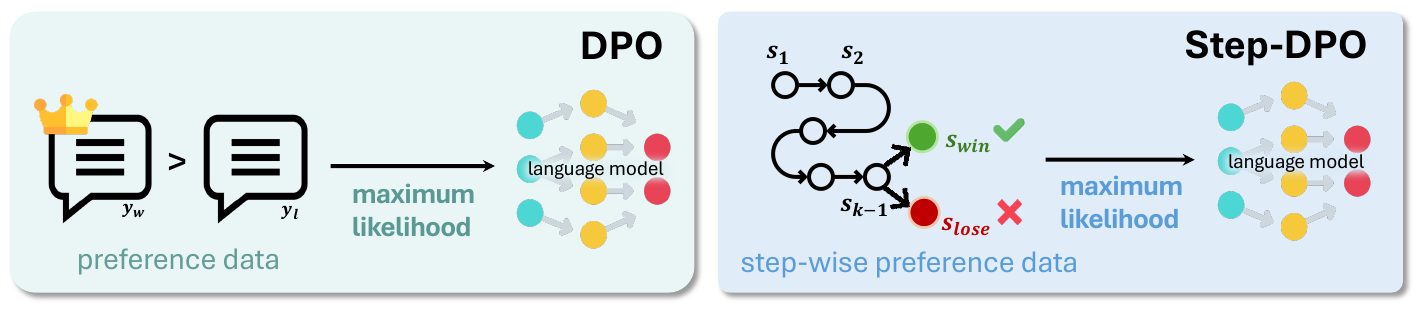}
\end{center}
\vspace{-0.4cm}
\caption{Comparison between DPO and Step-DPO.}
\label{fig:dpo_vs_step-dpo}
\end{figure}

In this work, we introduce Step-DPO, where each intermediate reasoning step is treated as the basic unit for preference optimization. As illustrated in Fig.~\ref{fig:dpo_vs_step-dpo}, unlike vanilla DPO, which only considers preference optimization between complete answers (i.e., $p(y_{win}|x)$ and $p(y_{lose}|x)$), Step-DPO examines the step-by-step answer (i.e., $y=s_1, ..., s_n$) and specifically targets the first erroneous reasoning step. Step-DPO aims to select a correct reasoning step and reject an incorrect one, given a math problem and several initial correct reasoning steps (i.e., maximize $p(s_{win}|x;s_1, s_2, ..., s_{k-1})$ and minimize $p(s_{lose}|x;s_1, s_2, ..., s_{k-1})$). This transition allows the model to easily locate erroneous tokens for effective optimization, significantly enhancing long-chain reasoning.

Moreover, we present an effective and economical pipeline to collect pair-wise preference data, resulting in a high-quality dataset for Step-DPO. This dataset contains approximately 10K samples, each consisting of: 1) a mathematical problem, 2) prior reasoning steps, 3) the chosen step, and 4) the rejected step. Our three-step pipeline for dataset construction includes: 1) Error collection, 2) Step localization, and 3) Rectification. Notably, the chosen reasoning step is generated by the model itself, as we find that in-distribution data (i.e., self-generated data) is more effective than out-of-distribution data (e.g., data written by humans or GPT-4) for Step-DPO, as shown in Table~\ref{table:ablation_data}.

With this curated dataset, mathematical reasoning performance can be significantly boosted with only hundreds of training steps, as demonstrated in Fig.~\ref{fig:acc_scatter}. For instance, fine-tuning Qwen-72B-Instruct with Step-DPO results in a model achieving 70.8\% accuracy on MATH and 94.0\% on GSM8K, surpassing a series of closed-source models, including GPT-4-1106, Claude-3-Opus, and Gemini-1.5-Pro.

\section{Related Works}

\subsection{Mathematical Reasoning}

Large Language Models (LLMs) have exhibited substantial reasoning capabilities, primarily due to their auto-regressive nature, which allows them to predict the next token based on contextual information. However, these models still struggle with long-chain reasoning tasks, particularly in mathematical contexts. Several prior studies~\citep{yao2024tree,chen2024alphamath, yoran2023answering, li2023camel, tong2024can, fu2022complexity, zhou2022least} have attempted to enhance the Chain-of-Thought (CoT) inference framework~\citep{wei2022chain} to address this issue. While these efforts have led to significant improvements in certain tasks, they have not fully mitigated common hallucinations and have limited generalizability across all reasoning tasks.

Another research direction~\citep{yu2023metamath, luo2023wizardmath, yue2023mammoth, liu2024augmenting, lu2024mathgenie, xu2024chatglm, li2024common, shao2024deepseekmath, xin2024deepseek, zhou2024jiuzhang3, liu2023improving, ying2024internlm, yue2024mammoth2, tang2024mathscale, mitra2024orca, yuan2023scaling} focuses on various data augmentation techniques, such as rephrasing, extension, and evolution, for supervised fine-tuning (SFT). These methods have significantly enhanced the reasoning abilities of LLMs, but their performance plateaus once the data reaches a certain volume. Additionally, methods like those proposed by \cite{wang2023mathcoder, liao2024mario, toshniwal2024openmathinstruct, gou2023tora} employ external tools, such as Python, to substantially reduce calculation errors. 

Other approaches~\citep{azerbayev2023llemma, shao2024deepseekmath, lin2024rho, ying2024internlm, wang2023generative} involve continued pre-training on extensive, high-quality math-related datasets, which markedly improve mathematical reasoning capabilities. Recent studies~\citep{xu2024chatglm, ying2024internlm} have explored reinforcement learning to mitigate hallucinations in mathematical reasoning. Works like \cite{lightman2023let, shao2024deepseekmath, wang2023math} emphasize the importance of step-by-step verification in reinforcement learning for mathematical problems. However, these methods still rely on the quality of the reward model and require the complex training pipelines of RLHF. Building on this line of research, we propose Step-DPO, a simpler, more effective, and more efficient method.

\subsection{Reinforcement Learning from Human Feedback}

Supervised fine-tuning (SFT) can align models with human preferences. However, as the probability of preferred outputs increases, so does the likelihood of undesirable ones, leading to hallucinations. To generate more reliable outputs, Reinforcement Learning from Human Feedback (RLHF)~\citep{christiano2017deep, ouyang2022training} has been introduced for LLM alignment. This approach involves training a reward model with comparison data and then using this reward model to optimize the policy model. The final performance heavily depends on the quality of the reward model, and the training pipeline is quite complex.

To simplify this process, Direct Preference Optimization (DPO)~\citep{rafailov2024direct} was proposed, which directly uses pair-wise preference data for model optimization. This transition significantly streamlines the training pipeline. While DPO has proven effective in chat benchmarks, it offers only marginal benefits for mathematical reasoning. Inheriting the principles of DPO, Step-DPO is specifically designed for long-chain reasoning and has shown significant performance improvements in solving math word problems.

\section{Step-DPO}

In this section, we elaborate on the proposed Step-DPO. First, we present step-wise formulation in Sec.~\ref{sec:step_dpo}, a novel approach designed to enhance long-chain reasoning abilities by building on DPO. Next, in Sec.~\ref{sec:data_const}, we illustrate a pipeline for constructing the step-wise preference dataset for Step-DPO. Both components are essential for achieving the desired performance improvements.

\subsection{Step-wise Formulation}
\label{sec:step_dpo}

\paragraph{Preliminary.} Reinforcement Learning from Human Feedback (RLHF)~\citep{christiano2017deep} is an effective approach for enhancing the robustness, factuality, and safety of LLMs~\citep{ouyang2022training}. RLHF consists of two training phases: 1) reward model training, and 2) policy model training. However, the final performance of RLHF is highly sensitive to various hyperparameters in both phases, necessitating meticulous tuning.

To avoid this complex training pipeline, \cite{rafailov2024direct} proposed Direct Preference Optimization (DPO), which directly uses pair-wise preference data to optimize the policy model with an equivalent optimization objective. Specifically, given an input prompt \(x\), and a preference data pair \((y_{win}, y_{lose})\), DPO aims to maximize the probability of the preferred output \(y_{win}\) and minimize that of the undesirable output \(y_{lose}\). The optimization objective is formulated as:
\begin{align}
\begin{aligned}
    \mathcal{L}_{DPO}(\theta) = -\mathbb{E}_{(x,y_{win},y_{lose}) \sim D} [\log \sigma (\beta \log \frac{\pi_{\theta}(y_{win}|x)}{\pi_{ref}(y_{win}|x)} - \beta \log \frac{\pi_{\theta}(y_{lose}|x)}{\pi_{ref}(y_{lose}|x)})],
\end{aligned}
\end{align}
where \(D\) is the pair-wise preference dataset, \(\sigma\) is the sigmoid function, \(\pi_{\theta}(\cdot|x)\) is the policy model to be optimized, \(\pi_{ref}(\cdot|x)\) is the reference model kept unchanged during training, and the hyperparameter \(\beta\) controls the distance from the reference model.

\paragraph{Our Solution.} While DPO has proven effective in chat benchmarks, it brings only marginal improvements for long-chain reasoning tasks such as mathematical problems, as shown in Fig.~\ref{fig:line} and Table~\ref{table:ablation_dpo}. This limitation arises because most undesirable answers in these tasks do not contain errors initially; the first error often appears midway through the reasoning process. Rejecting an entire undesirable answer in DPO may also discard preceding correct reasoning steps, introducing significant noise and negatively impacting training.

Analogous to how teachers correct students by pinpointing specific errors rather than dismissing entire answers, our proposed Step-DPO provides more detailed supervision by identifying the specific erroneous reasoning step. This granular focus allows the model to swiftly locate, rectify, and avoid erroneous steps.

Specifically, the answer \(y\) can be decomposed into a sequence of reasoning steps \(y = s_1, \ldots, s_n\), where \(s_i\) is the \(i\)-th reasoning step. As illustrated in Fig.~\ref{fig:dpo_vs_step-dpo}, given a prompt \(x\) and a series of initial correct reasoning steps \(s_{1 \sim k-1} = s_1, \ldots, s_{k-1}\), Step-DPO aims to maximize the probability of the correct next reasoning step \(s_{win}\) and minimize the probability of the incorrect one \(s_{lose}\). This objective can be formulated as:
\begin{align}
\label{eq:loss_fn}
\begin{aligned}
    \mathcal{L}(\theta) = -\mathbb{E}_{(x,s_{1\sim k-1},s_{win},s_{lose}) \sim D}[\log \sigma (\beta \log \frac{\pi_{\theta}(s_{win}|x;s_{1\sim k-1})}{\pi_{ref}(s_{win}|x;s_{1\sim k-1})} - \beta \log \frac{\pi_{\theta}(s_{lose}|x;s_{1\sim k-1})}{\pi_{ref}(s_{lose}|x;s_{1\sim k-1})})].
\end{aligned}
\end{align}

\subsection{In-distribution Data Construction}
\label{sec:data_const}

\begin{figure}
\begin{center}
\includegraphics[width=1.0\linewidth]{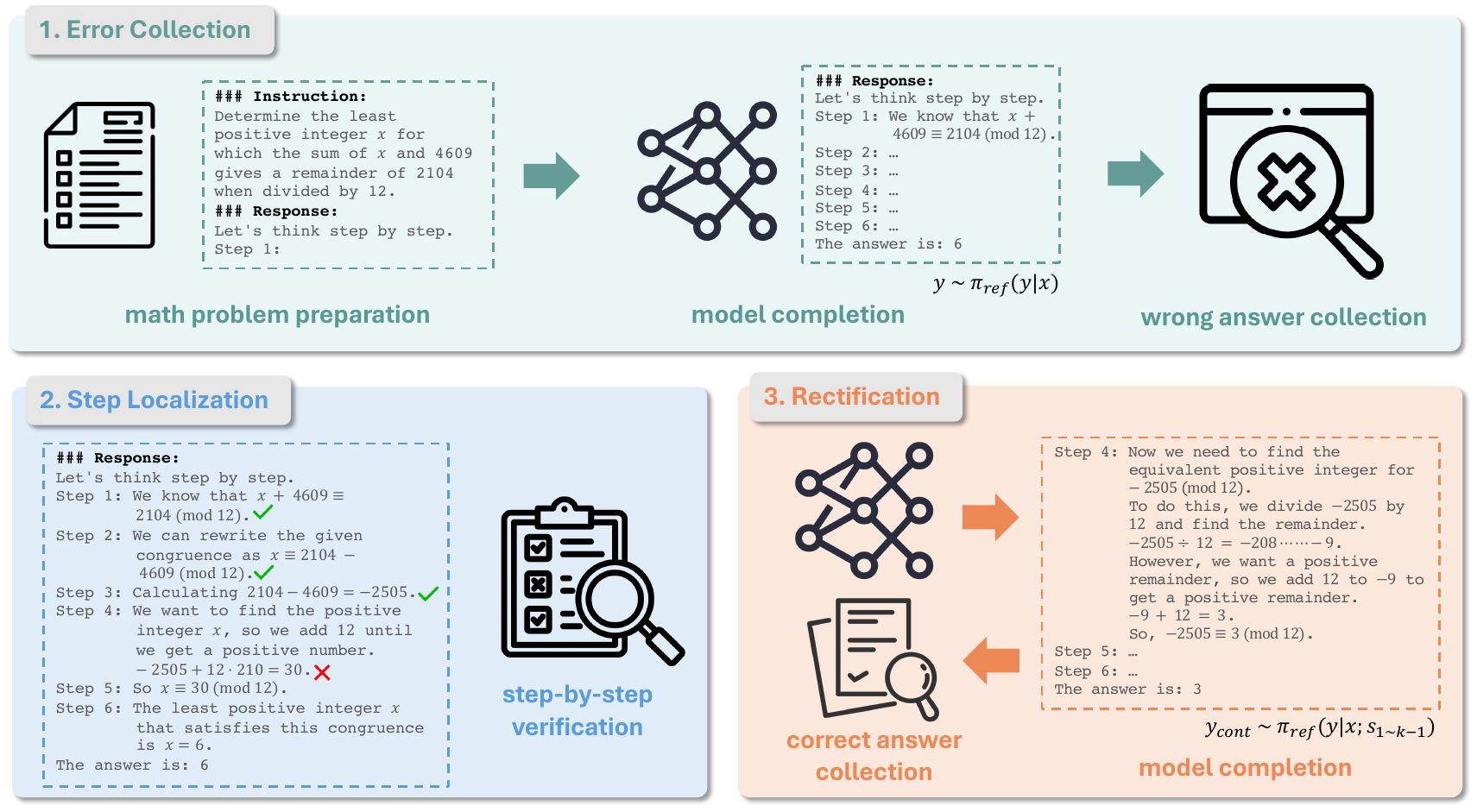}
\end{center}
\vspace{-0.4cm}
\caption{Data construction pipeline for Step-DPO.}
\label{fig:data_collection}
\end{figure}

\begin{figure}[t]
\begin{center}
\includegraphics[width=0.9\linewidth]{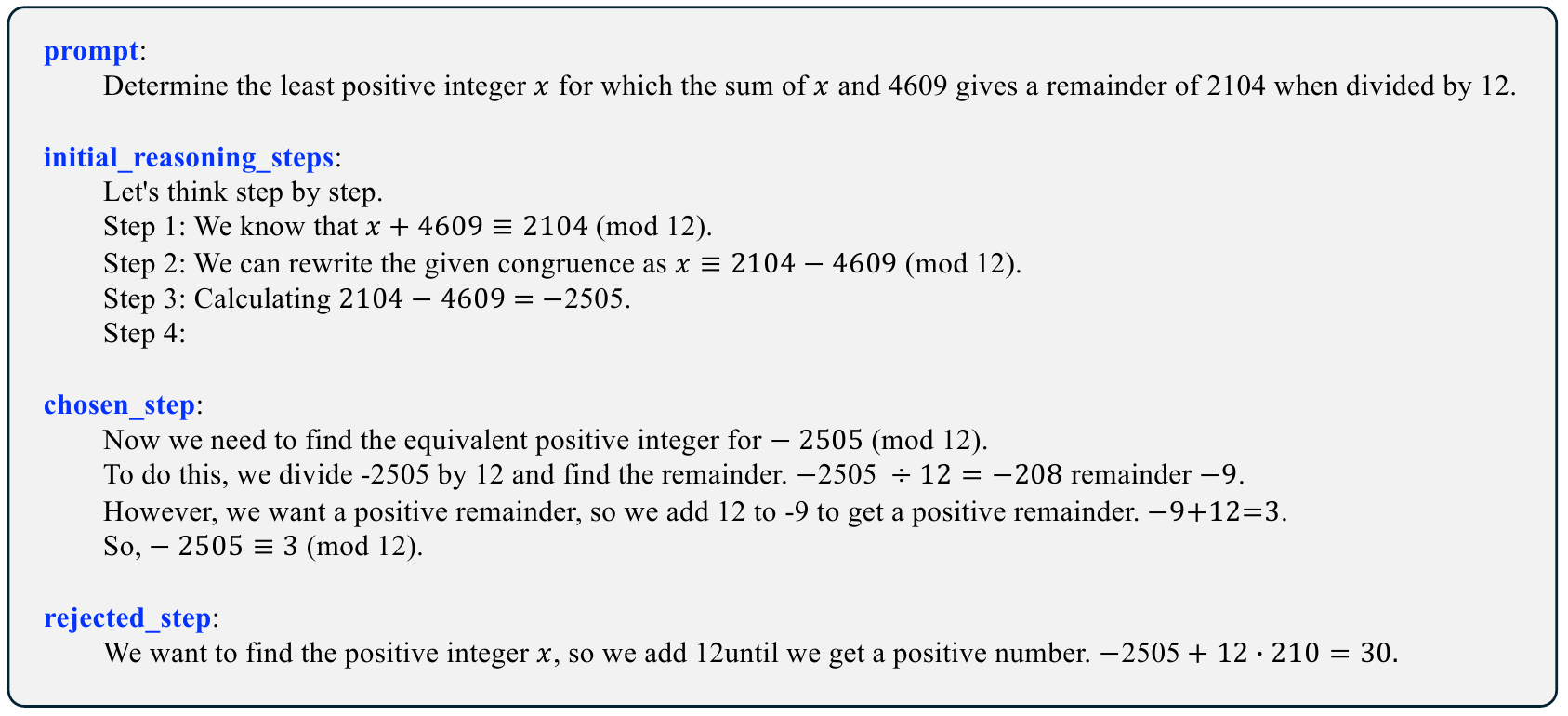}
\end{center}
\vspace{-0.4cm}
\caption{An example of preference data sample for Step-DPO.}
\label{fig:example}
\end{figure}

According to the optimization target of Step-DPO, we need to create a corresponding high-quality pair-wise preference dataset. Each data sample should comprise four entries: 1) prompt \( x \); 2) initial reasoning steps \( s_{1 \sim k-1} = s_1, \ldots, s_{k-1} \); 3) preferred reasoning step \( s_{win} \); 4) undesirable reasoning step \( s_{lose} \), as shown in Fig.~\ref{fig:example}. To obtain a high-quality dataset, we propose a data construction pipeline illustrated in Fig.~\ref{fig:data_collection}, which includes the following three steps.

\paragraph{Error collection.} First, we collect a set \( D_0 = \{(x, \hat{y})\} \) of mathematical problems \( x \) with ground-truth answers \( \hat{y} \). Each mathematical problem \( x \) is then used as a prompt to infer answers using the initial model \( \pi_{ref} \). Before inference, we add the step-wise Chain-of-Thought (CoT) prefix for prompting, i.e., \texttt{"Let's think step by step. Step 1:"}. This ensures that the model’s inference results are structured into multiple reasoning steps, with each step explicitly starting with \texttt{"Step i:"}. Upon completion of inference, we obtain the model answers \( y \) for each mathematical problem \( x \). We then select instances where the final answer \( y \) differs from the ground truth \( \hat{y} \), resulting in a dataset of erroneous inference results, denoted as \( D_1 = \{(x, \hat{y}, y) | x \in D_0\} \).

\paragraph{Step localization.} Given that each erroneous inference result is explicitly presented as a sequence of reasoning steps \( y = s_1, s_2, \ldots, s_n \), we proceed to verify the correctness of each reasoning step until we find the first error and record its step number \( k \). This process can be done manually or using GPT-4. We select \( s_k \) as the erroneous reasoning step \( s_{lose} \), resulting in a dataset that contains the erroneous steps, denoted as \( D_2 = \{(x, \hat{y}, s_{1 \sim k-1}, s_{lose}) | x \in D_1\} \).

\paragraph{Rectification.} To obtain the corresponding correct reasoning step for each sample in \( D_2 \), we need to sample multiple outputs \( y_{cont} \) by inferring the model \( \pi_{ref} \) with the prompt \( x \) and the preceding correct reasoning steps \( s_{1 \sim k-1} \). This process is formulated as:
\begin{align}
\begin{aligned}
     y_{cont} \sim \pi_{ref}(y|x;s_{1\sim k-1}).
\end{aligned}
\end{align}
We retain those outputs where the final answer matches the ground truth. Among the remaining outputs, we select the first reasoning step in \( y_{cont} \) as \( s_{win} \), resulting in the final dataset \( D = \{(x, s_{1 \sim k-1}, s_{lose}, s_{win}) | x \in D_2\} \). An example of a resulting data sample is shown in Fig.~\ref{fig:example}. 

Notably, some cases may have correct final answers but erroneous intermediate reasoning steps. Therefore, we may need to further filter out samples where \( s_{win} \) is incorrect, which can be done manually or by GPT-4. We omit this process in the notations for simplicity, and more details are provided in the appendix.

It is important to note that the data pipeline is user-friendly. In this data pipeline, humans or GPT-4 are only required to locate errors and rank answers, and they do not need to write answers or rectifications by themselves. 

\label{sec:in_dist_data}
We also note that the use of in-distribution data is crucial. When selecting \( s_{win} \), we use outputs generated by the model \( \pi_{ref} \) rather than answers rectified by humans or GPT-4. Since human or GPT-4 rectified answers \( s_{win}^{ood} \) are out-of-distribution (OOD) regarding the model \( \pi_{ref} \), the log-probability of outputting \( s_{win}^{ood} \) (i.e., \( \log \pi_{ref}(s_{win}^{ood} | x) \)) is significantly lower than that of an in-distribution (ID) output \( \log \pi_{ref}(s_{win}^{id} | x) \). Moreover, it is challenging for the policy model \( \pi_{\theta} \) to learn to increase the probability of \( s_{win}^{ood} \) due to gradient decay issues (detailed in the appendix). Consequently, adopting self-generated in-distribution data as the preferred answer proves to be a more effective way of aligning with human preferences.

\section{Experiments}

In this section, we first introduce the experimental setup in Sec.~\ref{sec:exp_setup}. Then, we present the main results in Sec.~\ref{sec:result}, which include an exhaustive performance comparison. Moreover, we conduct an extensive ablation study in Sec.~\ref{sec:ablation}. Finally, a few demonstrations are shown in Sec.~\ref{sec:demo} to further understand Step-DPO.

\subsection{Experimental Setup}
\label{sec:exp_setup}

\paragraph{Network Architecture.}
Our experiments are based on various base models, including the Qwen2 and Qwen1.5 series~\citep{qwen}, Meta-Llama-3-70B~\citep{touvron2023llama}, and deepseek-math-7b-base~\citep{shao2024deepseekmath}.

\begin{table}[p]
    \centering
    \caption{Math reasoning performance comparison on MATH and GSM8K across various models. general: general-purpose model. open: open-source.}
    \label{table:exp_comp}
    \vspace{0.1cm}
    \tabcolsep=0.06cm
    {
        \begin{threeparttable}
        \begin{tabular}{ l | c | c | c | c c}
            \toprule
            
            Model & size & general & open & MATH (\%) & GSM8K (\%) \\

            \specialrule{0em}{0pt}{1pt}
            \hline
            \specialrule{0em}{0pt}{1pt}

            GPT-3.5-Turbo & - & \Checkmark & \XSolidBrush & 42.5 & 92.0 \\

            Gemini-1.5-Pro (Feb)~\citep{reid2024gemini} & - & \Checkmark & \XSolidBrush & 58.5 & 91.7 \\

            Gemini-1.5-Pro (May)~\citep{reid2024gemini} & - & \Checkmark & \XSolidBrush & 67.7 & 90.8 \\
            
            Claude-3-Opus & - & \Checkmark & \XSolidBrush & 60.1 & 95.0 \\
            
            GPT-4-1106~\citep{achiam2023gpt} & - & \Checkmark & \XSolidBrush & 64.3 & 91.4 \\
            
            GPT-4-Turbo-0409~\citep{achiam2023gpt} & - & \Checkmark & \XSolidBrush & 73.4 & 93.7 \\
            
            GPT-4o-0513 & - & \Checkmark & \XSolidBrush & 76.6 & 95.8 \\

            \specialrule{0em}{0pt}{1pt}
            \hline
            \specialrule{0em}{0pt}{1pt}


            Llama-3-8B-Instruct~\citep{touvron2023llama} & 8B & \Checkmark & \Checkmark & 30.0 & 79.6 \\

            Qwen2-7B-Instruct~\citep{qwen} & 7B & \Checkmark & \Checkmark & 49.6 & 82.3 \\
            
            Llama-3-70B-Instruct~\citep{touvron2023llama} & 70B & \Checkmark & \Checkmark & 50.4 & 93.0 \\
            
            DeepSeek-Coder-V2-Instruct~\citep{zhu2024deepseek} & 236B & \XSolidBrush & \Checkmark & 75.7 & 94.9 \\
            
            \specialrule{0em}{0pt}{1pt}
            \hline
            \specialrule{0em}{0pt}{1pt}

            Code-Llama-7B~\citep{roziere2023code} & 7B & \XSolidBrush & \Checkmark & 13.0 & 25.2 \\
            
            MAmooTH-CoT~\citep{yue2023mammoth} & 7B & \XSolidBrush & \Checkmark & 10.4 & 50.5 \\
            
            WizardMath~\citep{luo2023wizardmath} & 7B & \XSolidBrush & \Checkmark & 10.7 & 54.9 \\
            
            MetaMath~\citep{yu2023metamath} & 7B & \XSolidBrush & \Checkmark & 19.8 & 66.5 \\

            MetaMath-Mistral-7B~\citep{yu2023metamath} & 7B & \XSolidBrush & \Checkmark & 28.2 & 77.7 \\

            MathScale-Mistral~\cite{tang2024mathscale} & 7B & \XSolidBrush & \Checkmark & 35.2 & 74.8 \\

            InternLM-Math-7B~\citep{ying2024internlm} & 7B & \XSolidBrush & \Checkmark & 34.6 & 78.1 \\

            Xwin-Math-Mistral-7B~\citep{li2024common} & 7B & \XSolidBrush & \Checkmark & 43.7 & 89.2 \\

            MAmmoTH2-7B-Plus~\citep{yue2024mammoth2} & 7B & \XSolidBrush & \Checkmark & 45.0 & 84.7 \\

            MathGenieLM-Mistral~\citep{lu2024mathgenie} & 7B & \XSolidBrush & \Checkmark & 45.1 & 80.5 \\

            InternLM-Math-20B~\citep{ying2024internlm} & 20B & \XSolidBrush & \Checkmark & 37.7 & 82.6 \\
            
            MathGenieLM-InternLM2~\citep{lu2024mathgenie} & 20B & \XSolidBrush & \Checkmark & 55.7 & 87.7 \\

            \specialrule{0em}{0pt}{1pt}
            \hline
            \specialrule{0em}{0pt}{1pt}

            DeepSeekMath-Instruct~\citep{shao2024deepseekmath} & 7B & \XSolidBrush & \Checkmark & 46.8 & 82.9 \\
            
            DeepSeekMath-RL~\citep{shao2024deepseekmath} & 7B & \XSolidBrush & \Checkmark & 51.7 & 88.2 \\
            
            \rowcolor{Gray}DeepSeekMath-RL + Step-DPO & 7B & \XSolidBrush & \Checkmark & 53.2 \textcolor[RGB]{20,160,20}{(+1.5)} & 88.7 \textcolor[RGB]{20,160,20}{(+0.5)} \\
            
            \specialrule{0em}{0pt}{1pt}
            \hline
            \specialrule{0em}{0pt}{1pt}

            DeepSeekMath-Base-SFT$^{\dagger}$ & 7B & \XSolidBrush & \Checkmark & 52.9 & 86.7 \\
            
            \rowcolor{Gray}DeepSeekMath-Base-SFT + Step-DPO & 7B & \XSolidBrush & \Checkmark & 54.1 \textcolor[RGB]{20,160,20}{(+1.2)} & 86.7 \textcolor[RGB]{20,160,20}{(+0.0)} \\
            
            \specialrule{0em}{0pt}{1pt}
            \hline
            \specialrule{0em}{0pt}{1pt}

            Qwen2-7B-SFT$^{\dagger}$ & 7B & \XSolidBrush & \Checkmark & 54.8 & 88.2 \\
            
            \rowcolor{Gray}Qwen2-7B-SFT + Step-DPO & 7B & \XSolidBrush & \Checkmark & 55.8 \textcolor[RGB]{20,160,20}{(+1.0)} & 88.5 \textcolor[RGB]{20,160,20}{(+0.3)} \\
            
            \specialrule{0em}{0pt}{1pt}
            \hline
            \specialrule{0em}{0pt}{1pt}

            
            
            

            Qwen1.5-32B-SFT$^{\dagger}$ & 32B & \XSolidBrush & \Checkmark & 54.9 & 90.0 \\
            
            \rowcolor{Gray}Qwen1.5-32B-SFT + Step-DPO & 32B & \XSolidBrush & \Checkmark & 56.9 \textcolor[RGB]{20,160,20}{(+2.0)} & 90.9 \textcolor[RGB]{20,160,20}{(+0.9)} \\
            
            \specialrule{0em}{0pt}{1pt}
            \hline
            \specialrule{0em}{0pt}{1pt}

            Qwen2-57B-A14B-SFT$^{\dagger}$ & 57B & \XSolidBrush & \Checkmark & 54.6 & 89.8 \\
            
            \rowcolor{Gray}Qwen2-57B-A14B-SFT + Step-DPO & 57B & \XSolidBrush & \Checkmark & 56.5 \textcolor[RGB]{20,160,20}{(+1.9)} & 90.0 \textcolor[RGB]{20,160,20}{(+0.2)} \\
            
            \specialrule{0em}{0pt}{1pt}
            \hline
            \specialrule{0em}{0pt}{1pt}

            Llama-3-70B-SFT$^{\dagger}$ & 70B & \XSolidBrush & \Checkmark & 56.9 & 92.2 \\
            
            \rowcolor{Gray}Llama-3-70B-SFT + Step-DPO & 70B & \XSolidBrush & \Checkmark & 59.5 \textcolor[RGB]{20,160,20}{(+2.6)} & 93.3 \textcolor[RGB]{20,160,20}{(+1.1)} \\
            
            \specialrule{0em}{0pt}{1pt}
            \hline
            \specialrule{0em}{0pt}{1pt}

            Qwen2-72B-SFT$^{\dagger}$ & 72B & \XSolidBrush & \Checkmark & 61.7 & 92.9 \\
            
            \rowcolor{Gray}Qwen2-72B-SFT + Step-DPO & 72B & \XSolidBrush & \Checkmark & 64.7 \textcolor[RGB]{20,160,20}{(+3.0)} & 93.9 \textcolor[RGB]{20,160,20}{(+1.0)} \\

            \specialrule{0em}{0pt}{1pt}
            \hline
            \specialrule{0em}{0pt}{1pt}

            Qwen2-72B-Instruct~\citep{qwen} & 72B & \Checkmark & \Checkmark & 59.7 & 91.1 \\
            
            Qwen2-72B-Instruct $^{\ddagger}$ & 72B & \Checkmark & \Checkmark & 69.4 & 92.4 \\
            
            \rowcolor{Gray}Qwen2-72B-Instruct + Step-DPO $^{\ddagger}$ & 72B & \Checkmark & \Checkmark & \textbf{70.8} \textcolor[RGB]{20,160,20}{(+1.4)} & \textbf{94.0} \textcolor[RGB]{20,160,20}{(+1.6)} \\

            \bottomrule
        \end{tabular}
        \begin{tablenotes}
          \small
          \item $\;\;^{\dagger}$ Supervised fine-tuned models with our 299K SFT data based on the open-source base model.
          \item $\;\;^{\ddagger}$ Reproduced using our prompt
        \end{tablenotes}
        \end{threeparttable}
    }
\end{table}

\begin{table}[t]
    \centering
    \caption{Math reasoning performance comparison on compitition-level math problems, i.e., AIME 2024 and Odyssey-MATH. Note that the training data for Step-DPO is the same as before.}
    \label{table:competition_results}
    \vspace{0.3cm}
    \tabcolsep=0.05cm
    {
        \begin{threeparttable}
        \begin{tabular}{ l | c | c | c c}
            \toprule
            
            Model & size & open & AIME & Odyssey-MATH (\%) \\

            \specialrule{0em}{0pt}{1pt}
            \hline
            \specialrule{0em}{0pt}{1pt}

            Gemini-1.5-Pro~\citep{reid2024gemini} & - & \XSolidBrush & 2 / 30 & 45.0 \\
            
            Claude-3-Opus & - & \XSolidBrush & 2 / 30 & 40.6 \\
            
            GPT-4-1106~\citep{achiam2023gpt} & - & \XSolidBrush & 1 / 30 & 49.1 \\
            
            GPT-4-Turbo-0409~\citep{achiam2023gpt} & - & \XSolidBrush & 3 / 30 & 46.8  \\
            
            GPT-4o-0513 & - & \XSolidBrush & 2 / 30 & 53.2  \\

            \specialrule{0em}{0pt}{1pt}
            \hline
            \specialrule{0em}{0pt}{1pt}

            DeepSeek-Coder-V2-Lite-Instruct~\citep{zhu2024deepseek} & 16B & \Checkmark & 0 / 30 & 44.4  \\
            
            Llama-3-70B-Instruct~\citep{touvron2023llama} & 70B & \Checkmark & 1 / 30 & 27.9  \\
            
            DeepSeek-Coder-V2-Instruct~\citep{zhu2024deepseek} & 236B & \Checkmark & 4 / 30 & 53.7  \\


            
            
            \specialrule{0em}{0pt}{1pt}
            \hline
            \specialrule{0em}{0pt}{1pt}

            Qwen2-72B-SFT$^{\dagger}$ & 72B & \Checkmark & 1 / 30 & 44.2  \\
            
            \rowcolor{Gray}Qwen2-72B-SFT + Step-DPO & 72B & \Checkmark & 3 / 30 & 47.0 \textcolor[RGB]{20,160,20}{(+2.8)} \\
            
            \specialrule{0em}{0pt}{1pt}
            \hline
            \specialrule{0em}{0pt}{1pt}

            Qwen2-72B-Instruct~\citep{qwen} & 72B & \Checkmark & 5 / 30 & 47.0 \\
            
            \rowcolor{Gray}Qwen2-72B-Instruct + Step-DPO & 72B & \Checkmark & 4 / 30 & 50.1 \textcolor[RGB]{20,160,20}{(+3.1)} \\
            
            \bottomrule
        \end{tabular}
        \begin{tablenotes}
          \small
          \item $\;\;^{\dagger}$ Supervised fine-tuned models with our 299K SFT data based on the open-source base model.
        \end{tablenotes}
        \end{threeparttable}
    }
\end{table}

\paragraph{Datasets.} 
\label{sec:exp_data}
In supervised fine-tuning (SFT), we use augmented mathematical problems from MetaMath~\citep{yu2023metamath} and MMIQC~\citep{liu2024augmenting} to infer step-by-step responses with DeepSeekMath, as the SFT data used in DeepSeekMath~\citep{shao2024deepseekmath} is not publicly available. After filtering out responses with erroneous final answers, we obtain 374K SFT data. Of these, 299K are used for SFT, and the remainder is used for further Step-DPO training.

In the Step-DPO phase, alongside the remaining SFT data, we also incorporate a subset of AQuA~\citep{ling2017program}. These datasets are processed as described in Sec.~\ref{sec:data_const}, resulting in 10K pair-wise preference data for Step-DPO.

For evaluation, we use the widely adopted MATH~\citep{hendrycks2021measuring} and GSM8K~\citep{cobbe2021training} datasets. Accuracy in these datasets serves as the evaluation metric. The MATH test set contains 5000 mathematical problems spanning 5 difficulty levels and 7 subjects, including algebra, counting and probability, geometry, intermediate algebra, number theory, prealgebra, and precalculus. The GSM8K test set includes 1319 mathematical problems, each with a step-by-step solution and a ground-truth answer. The problems in GSM8K are generally easier than those in MATH. Besides, we also use completition-level problems in American Invitational Mathematics Examination (AIME)~\citep{AIME} and Odyssey-MATH~\citep{netmindmath} to evaluate the math reasoning capabilities in solving hard problems.

\paragraph{Implementation Details.}
First, we use the 299K SFT data for supervised fine-tuning on the base models, obtaining the SFT models. We train 7B models for 3 epochs and models larger than 30B for 2 epochs. The global batch size is set to 256, and the learning rate is set to 5e-6. We use the AdamW optimizer with a linear decay learning rate scheduler, setting the warmup ratio to 0.03. DeepSpeed ZeRO3 with CPU offload is used to reduce GPU memory usage during training.

Next, we perform Step-DPO based on the SFT models. For Step-DPO, we train 7B models for 8 epochs and models larger than 30B for 4 epochs. The global batch size is set to 128, and the learning rate is set to 5e-7. The hyperparameter \( \beta \) is set to 0.5 for the 72B model and 0.4 for others. We use the AdamW optimizer and a cosine learning rate scheduler, with the warmup ratio set to 0.1.

\subsection{Results}
\label{sec:result}

\begin{table}[t]
    \centering
    \tabcolsep=0.16cm
    \caption{Performance comparison between DPO and Step-DPO. We use only 5K data for training in this ablation study.}
    \vspace{0.2cm}
    \label{table:ablation_dpo}   
    {
        \begin{footnotesize}
        \begin{tabular}{ c | c |  c | c }
            \toprule
            \textbf{Model} & Qwen2-7B-SFT & Qwen2-7B-SFT + DPO (5K) & Qwen2-7B-SFT + Step-DPO (5K) \\

            \specialrule{0em}{0pt}{1pt}
            \hline
            \specialrule{0em}{0pt}{1pt}
            
            \textbf{MATH} (\%) & 54.8 & 55.0 & \textbf{55.8} \\

            \specialrule{0em}{0pt}{1pt}
            \hline
            \specialrule{0em}{0pt}{1pt}
            
            \textbf{Model} & Qwen2-72B-SFT & Qwen2-72B-SFT + DPO (5K) & Qwen2-72B-SFT + Step-DPO (5K) \\

            \specialrule{0em}{0pt}{1pt}
            \hline
            \specialrule{0em}{0pt}{1pt}
            
            \textbf{MATH} (\%) & 61.7 & 62.5 & \textbf{64.1} \\
            
            \bottomrule                                   
        \end{tabular}
        \end{footnotesize}
    }    
\end{table}

\begin{table}[t]
    \centering
    \tabcolsep=0.1cm
    \caption{Performance comparison between out-of-distribution and in-distribution data. \textbf{OOD}: out-of-distribution data. \textbf{ID}: in-distribution data.}
    \vspace{0.2cm}
    \label{table:ablation_data}   
    {
        \begin{footnotesize}
        \begin{tabular}{ c | c |  c | c }
            \toprule
            \textbf{Model} & Qwen2-7B-SFT & Qwen2-7B-SFT + Step-DPO (OOD) & Qwen2-7B-SFT + Step-DPO (ID)\\

            \specialrule{0em}{0pt}{1pt}
            \hline
            \specialrule{0em}{0pt}{1pt}
            
            \textbf{MATH} (\%) & 54.8 & 55.1 & \textbf{55.8} \\

            \bottomrule
        \end{tabular}
        \end{footnotesize}
    }    
\end{table}

\paragraph{Applying on open-source instruct models.} Table~\ref{table:exp_comp} presents a comprehensive comparison of various models, encompassing both open-source and closed-source models. Notably, Step-DPO can be directly integrated into open-source instruction models, such as DeepSeekMath-RL and Qwen2-72B-Instruct, leading to significant performance enhancements even after their prior RLHF training phase. This indicates that Step-DPO complements RLHF effectively. Specifically, when applied to Qwen2-72B-Instruct, Step-DPO achieves scores of 70.8\% and 94.0\% on the MATH and GSM8K test sets, respectively, surpassing a series of closed-source models, including GPT-4-1106, Claude-3-Opus, and Gemini-1.5-Pro.

\paragraph{Applying on SFT models.} To further substantiate the efficacy of Step-DPO, we applied it to SFT models. Initially, we performed supervised fine-tuning on the 299K SFT dataset mentioned in Sec.~\ref{sec:exp_data}, resulting in models such as DeepSeekMath-Base-SFT, Qwen2-7B-SFT, Qwen1.5-32B-SFT, Llama3-70B-SFT, and Qwen2-72B-SFT. Step-DPO proved highly effective, yielding significant improvements across various model sizes. Particularly, for models exceeding 70B parameters (i.e., Llama-3-70B-SFT and Qwen-2-72B-SFT), Step-DPO achieved approximately a 3\% performance boost on the MATH test set.

Interestingly, larger models exhibited greater performance gains from Step-DPO. We hypothesize that larger models have untapped potential that Step-DPO can exploit. If the performance ceiling is not reached through supervised fine-tuning (SFT), Step-DPO can help models approach their optimal performance.

\paragraph{Results on math competition problems.} To further illustrate the superiority of Step-DPO in mathematical reasoning, we evaluated the models on competition-level math problems, specifically AIME 2024 and Odyssey-MATH, as shown in Fig.~\ref{table:competition_results}. Despite the increased difficulty of these problems compared to MATH and GSM8K, Step-DPO significantly enhanced performance. On Odyssey-MATH, Step-DPO applied to Qwen2-72B-Instruct achieved 50.1\% accuracy, narrowing the performance gap with GPT-4o.

Notably, the models used the same Step-DPO training data for these competition-level problems as for problems of normal difficulty, highlighting Step-DPO's robust generalization capability.

\subsection{Ablation Study}
\label{sec:ablation}
To validate the effectiveness of Step-DPO and its data construction process, we conducted an extensive ablation study.

\paragraph{DPO vs. Step-DPO.} As discussed in Sec.~\ref{sec:step_dpo}, models utilizing vanilla DPO struggle to accurately identify errors in incorrect answers, providing only marginal benefits to mathematical reasoning performance. To verify this, we compared vanilla DPO and Step-DPO in terms of both accuracy in judging preferred versus undesirable outputs (left side of Fig.~\ref{fig:line}) and the reward margin between them (right side of Fig.~\ref{fig:line}). We also reported the final mathematical reasoning performance on the MATH test set in Table~\ref{table:ablation_dpo}. The results indicated that the benefits of DPO are limited and significantly less than those of Step-DPO. In this experiment, we used only 5K Step-DPO training data and 3K for evaluation in Fig.~\ref{fig:line}.

\paragraph{Out-of-Distribution vs. In-Distribution Data.}
\label{sec:ablation_data} The importance of in-distribution data was emphasized in Sec.~\ref{sec:in_dist_data}. To illustrate its effectiveness, we compared out-of-distribution and in-distribution data in Table~\ref{table:ablation_data}. Out-of-distribution data was generated by using GPT-4 to correct erroneous reasoning steps in incorrect answers, whereas in-distribution data was generated through the pipeline described in Sec.~\ref{sec:data_const}. The results in Table~\ref{table:ablation_data} underscore the critical role of in-distribution data in enhancing performance.

\subsection{Demonstrations}
\label{sec:demo}

\begin{figure}[t]
\begin{center}
\includegraphics[width=1.0\linewidth]{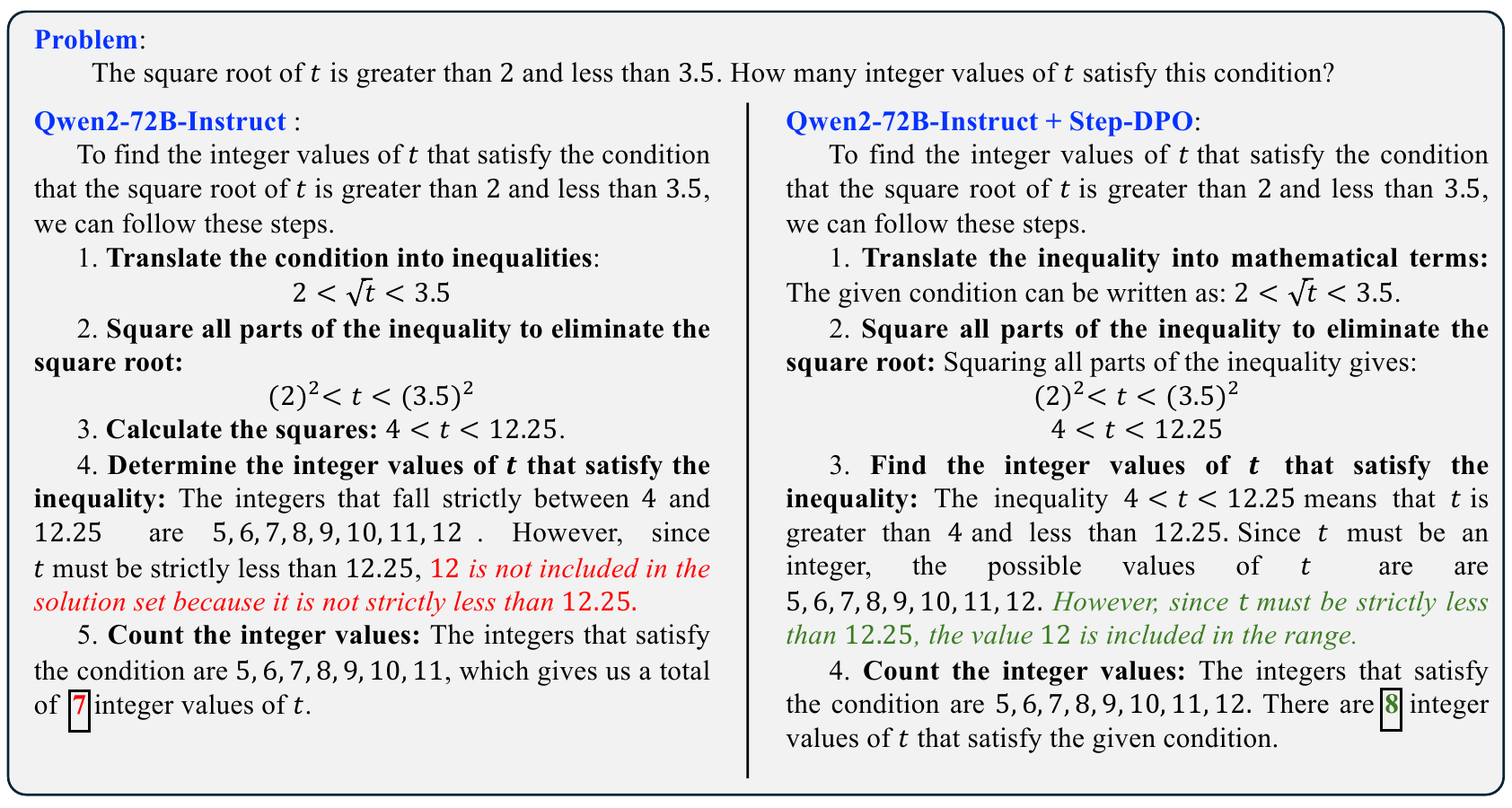}
\end{center}
\vspace{-0.4cm}
\caption{An example of comparison between Qwen2-72B-Instruct and Qwen2-72B-Instruct-Step-DPO.}
\label{fig:acc_scatter}
\end{figure}

As shown in Fig.~\ref{fig:acc_scatter}, we demonstrate an example of comparison between Qwen2-72B-Instruct and Qwen2-72B-Instruct-Step-DPO. It turns out that Step-DPO does well in correcting minor mistakes in previous models. More comparisons are provided in the appendix.

\section{Conclusion}

In this work, we proposed a simple, effective, and data-efficient method called Step-DPO. Unlike DPO, which compares preferences between holistic answers, Step-DPO uses a single reasoning step as the fundamental unit for preference comparison. This transition enables fine-grained process supervision for LLMs, facilitating the quick localization of errors within incorrect answers. Additionally, we introduced a data construction pipeline for Step-DPO, creating a dataset with 10K preference data pairs. Our results demonstrate the significant improvements achieved by Step-DPO and the 10K dataset, particularly for large models. We hope that Step-DPO will provide new insights into model alignment for long-chain reasoning problems.


\bibliography{iclr2024_conference}

\begin{thebibliography}{50}
\providecommand{\natexlab}[1]{#1}
\providecommand{\url}[1]{\texttt{#1}}
\expandafter\ifx\csname urlstyle\endcsname\relax
  \providecommand{\doi}[1]{doi: #1}\else
  \providecommand{\doi}{doi: \begingroup \urlstyle{rm}\Url}\fi

\bibitem[Achiam et~al.(2023)Achiam, Adler, Agarwal, Ahmad, Akkaya, Aleman, Almeida, Altenschmidt, Altman, Anadkat, et~al.]{achiam2023gpt}
Josh Achiam, Steven Adler, Sandhini Agarwal, Lama Ahmad, Ilge Akkaya, Florencia~Leoni Aleman, Diogo Almeida, Janko Altenschmidt, Sam Altman, Shyamal Anadkat, et~al.
\newblock Gpt-4 technical report.
\newblock \emph{arXiv:2303.08774}, 2023.

\bibitem[Azerbayev et~al.(2023)Azerbayev, Schoelkopf, Paster, Santos, McAleer, Jiang, Deng, Biderman, and Welleck]{azerbayev2023llemma}
Zhangir Azerbayev, Hailey Schoelkopf, Keiran Paster, Marco~Dos Santos, Stephen McAleer, Albert~Q Jiang, Jia Deng, Stella Biderman, and Sean Welleck.
\newblock Llemma: An open language model for mathematics.
\newblock \emph{arXiv:2310.10631}, 2023.

\bibitem[Bai et~al.(2023)Bai, Bai, Chu, Cui, Dang, Deng, Fan, Ge, Han, Huang, Hui, Ji, Li, Lin, Lin, Liu, Liu, Lu, Lu, Ma, Men, Ren, Ren, Tan, Tan, Tu, Wang, Wang, Wang, Wu, Xu, Xu, Yang, Yang, Yang, Yang, Yao, Yu, Yuan, Yuan, Zhang, Zhang, Zhang, Zhang, Zhou, Zhou, Zhou, and Zhu]{qwen}
Jinze Bai, Shuai Bai, Yunfei Chu, Zeyu Cui, Kai Dang, Xiaodong Deng, Yang Fan, Wenbin Ge, Yu~Han, Fei Huang, Binyuan Hui, Luo Ji, Mei Li, Junyang Lin, Runji Lin, Dayiheng Liu, Gao Liu, Chengqiang Lu, Keming Lu, Jianxin Ma, Rui Men, Xingzhang Ren, Xuancheng Ren, Chuanqi Tan, Sinan Tan, Jianhong Tu, Peng Wang, Shijie Wang, Wei Wang, Shengguang Wu, Benfeng Xu, Jin Xu, An~Yang, Hao Yang, Jian Yang, Shusheng Yang, Yang Yao, Bowen Yu, Hongyi Yuan, Zheng Yuan, Jianwei Zhang, Xingxuan Zhang, Yichang Zhang, Zhenru Zhang, Chang Zhou, Jingren Zhou, Xiaohuan Zhou, and Tianhang Zhu.
\newblock Qwen technical report.
\newblock \emph{arXiv:2309.16609}, 2023.

\bibitem[Chen et~al.(2024)Chen, Liao, Li, and Fan]{chen2024alphamath}
Guoxin Chen, Minpeng Liao, Chengxi Li, and Kai Fan.
\newblock Alphamath almost zero: process supervision without process.
\newblock \emph{arXiv:2405.03553}, 2024.

\bibitem[Christiano et~al.(2017)Christiano, Leike, Brown, Martic, Legg, and Amodei]{christiano2017deep}
Paul~F Christiano, Jan Leike, Tom Brown, Miljan Martic, Shane Legg, and Dario Amodei.
\newblock Deep reinforcement learning from human preferences.
\newblock \emph{NeurIPS}, 2017.

\bibitem[Cobbe et~al.(2021)Cobbe, Kosaraju, Bavarian, Chen, Jun, Kaiser, Plappert, Tworek, Hilton, Nakano, et~al.]{cobbe2021training}
Karl Cobbe, Vineet Kosaraju, Mohammad Bavarian, Mark Chen, Heewoo Jun, Lukasz Kaiser, Matthias Plappert, Jerry Tworek, Jacob Hilton, Reiichiro Nakano, et~al.
\newblock Training verifiers to solve math word problems.
\newblock \emph{arXiv:2110.14168}, 2021.

\bibitem[Fu et~al.(2022)Fu, Peng, Sabharwal, Clark, and Khot]{fu2022complexity}
Yao Fu, Hao Peng, Ashish Sabharwal, Peter Clark, and Tushar Khot.
\newblock Complexity-based prompting for multi-step reasoning.
\newblock In \emph{ICLR}, 2022.

\bibitem[Gou et~al.(2023)Gou, Shao, Gong, Yang, Huang, Duan, Chen, et~al.]{gou2023tora}
Zhibin Gou, Zhihong Shao, Yeyun Gong, Yujiu Yang, Minlie Huang, Nan Duan, Weizhu Chen, et~al.
\newblock Tora: A tool-integrated reasoning agent for mathematical problem solving.
\newblock \emph{arXiv:2309.17452}, 2023.

\bibitem[Hendrycks et~al.(2021)Hendrycks, Burns, Kadavath, Arora, Basart, Tang, Song, and Steinhardt]{hendrycks2021measuring}
Dan Hendrycks, Collin Burns, Saurav Kadavath, Akul Arora, Steven Basart, Eric Tang, Dawn Song, and Jacob Steinhardt.
\newblock Measuring mathematical problem solving with the math dataset.
\newblock \emph{arXiv:2103.03874}, 2021.

\bibitem[Hong et~al.(2024)Hong, Lee, and Thorne]{hong2024orpo}
Jiwoo Hong, Noah Lee, and James Thorne.
\newblock Orpo: Monolithic preference optimization without reference model.
\newblock \emph{arXiv:2403.07691}, 2024.

\bibitem[Li et~al.(2024)Li, Wang, Hu, Wei, Zheng, Hu, Zhang, and Peng]{li2024common}
Chen Li, Weiqi Wang, Jingcheng Hu, Yixuan Wei, Nanning Zheng, Han Hu, Zheng Zhang, and Houwen Peng.
\newblock Common 7b language models already possess strong math capabilities.
\newblock \emph{arXiv:2403.04706}, 2024.

\bibitem[Li et~al.(2023)Li, Hammoud, Itani, Khizbullin, and Ghanem]{li2023camel}
Guohao Li, Hasan Hammoud, Hani Itani, Dmitrii Khizbullin, and Bernard Ghanem.
\newblock Camel: Communicative agents for" mind" exploration of large language model society.
\newblock \emph{NeurIPS}, 2023.

\bibitem[Liao et~al.(2024)Liao, Luo, Li, Wu, and Fan]{liao2024mario}
Minpeng Liao, Wei Luo, Chengxi Li, Jing Wu, and Kai Fan.
\newblock Mario: Math reasoning with code interpreter output--a reproducible pipeline.
\newblock \emph{arXiv:2401.08190}, 2024.

\bibitem[Lightman et~al.(2023)Lightman, Kosaraju, Burda, Edwards, Baker, Lee, Leike, Schulman, Sutskever, and Cobbe]{lightman2023let}
Hunter Lightman, Vineet Kosaraju, Yura Burda, Harri Edwards, Bowen Baker, Teddy Lee, Jan Leike, John Schulman, Ilya Sutskever, and Karl Cobbe.
\newblock Let's verify step by step.
\newblock \emph{arXiv:2305.20050}, 2023.

\bibitem[Lin et~al.(2024)Lin, Gou, Gong, Liu, Shen, Xu, Lin, Yang, Jiao, Duan, et~al.]{lin2024rho}
Zhenghao Lin, Zhibin Gou, Yeyun Gong, Xiao Liu, Yelong Shen, Ruochen Xu, Chen Lin, Yujiu Yang, Jian Jiao, Nan Duan, et~al.
\newblock Rho-1: Not all tokens are what you need.
\newblock \emph{arXiv:2404.07965}, 2024.

\bibitem[Ling et~al.(2017)Ling, Yogatama, Dyer, and Blunsom]{ling2017program}
Wang Ling, Dani Yogatama, Chris Dyer, and Phil Blunsom.
\newblock Program induction by rationale generation: Learning to solve and explain algebraic word problems.
\newblock \emph{arXiv:1705.04146}, 2017.

\bibitem[Liu \& Yao(2024)Liu and Yao]{liu2024augmenting}
Haoxiong Liu and Andrew Chi-Chih Yao.
\newblock Augmenting math word problems via iterative question composing.
\newblock \emph{arXiv:2401.09003}, 2024.

\bibitem[Liu et~al.(2023)Liu, Singh, Freeman, Co-Reyes, and Liu]{liu2023improving}
Yixin Liu, Avi Singh, C~Daniel Freeman, John~D Co-Reyes, and Peter~J Liu.
\newblock Improving large language model fine-tuning for solving math problems.
\newblock \emph{arXiv:2310.10047}, 2023.

\bibitem[Lu et~al.(2024)Lu, Zhou, Ren, Wang, Shi, Pan, Zhan, and Li]{lu2024mathgenie}
Zimu Lu, Aojun Zhou, Houxing Ren, Ke~Wang, Weikang Shi, Junting Pan, Mingjie Zhan, and Hongsheng Li.
\newblock Mathgenie: Generating synthetic data with question back-translation for enhancing mathematical reasoning of llms.
\newblock \emph{arXiv:2402.16352}, 2024.

\bibitem[Luo et~al.(2023)Luo, Sun, Xu, Zhao, Lou, Tao, Geng, Lin, Chen, and Zhang]{luo2023wizardmath}
Haipeng Luo, Qingfeng Sun, Can Xu, Pu~Zhao, Jianguang Lou, Chongyang Tao, Xiubo Geng, Qingwei Lin, Shifeng Chen, and Dongmei Zhang.
\newblock Wizardmath: Empowering mathematical reasoning for large language models via reinforced evol-instruct.
\newblock \emph{arXiv:2308.09583}, 2023.

\bibitem[MAA(2024)]{AIME}
MAA.
\newblock American invitational mathematics examination, 2024.
\newblock URL \url{https://maa.org/math-competitions/american-invitational-mathematics-examination-aime}.

\bibitem[Mitra et~al.(2024)Mitra, Khanpour, Rosset, and Awadallah]{mitra2024orca}
Arindam Mitra, Hamed Khanpour, Corby Rosset, and Ahmed Awadallah.
\newblock Orca-math: Unlocking the potential of slms in grade school math.
\newblock \emph{arXiv:2402.14830}, 2024.

\bibitem[Netmind.AI(2024)]{netmindmath}
Netmind.AI.
\newblock Odyssey-math.
\newblock \url{https://github.com/protagolabs/odyssey-math/tree/main}, 2024.
\newblock Accessed: April 22, 2024.

\bibitem[Ouyang et~al.(2022)Ouyang, Wu, Jiang, Almeida, Wainwright, Mishkin, Zhang, Agarwal, Slama, Ray, et~al.]{ouyang2022training}
Long Ouyang, Jeffrey Wu, Xu~Jiang, Diogo Almeida, Carroll Wainwright, Pamela Mishkin, Chong Zhang, Sandhini Agarwal, Katarina Slama, Alex Ray, et~al.
\newblock Training language models to follow instructions with human feedback.
\newblock \emph{NeurIPS}, 2022.

\bibitem[Rafailov et~al.(2024)Rafailov, Sharma, Mitchell, Manning, Ermon, and Finn]{rafailov2024direct}
Rafael Rafailov, Archit Sharma, Eric Mitchell, Christopher~D Manning, Stefano Ermon, and Chelsea Finn.
\newblock Direct preference optimization: Your language model is secretly a reward model.
\newblock \emph{NeurIPS}, 2024.

\bibitem[Reid et~al.(2024)Reid, Savinov, Teplyashin, Lepikhin, Lillicrap, Alayrac, Soricut, Lazaridou, Firat, Schrittwieser, et~al.]{reid2024gemini}
Machel Reid, Nikolay Savinov, Denis Teplyashin, Dmitry Lepikhin, Timothy Lillicrap, Jean-baptiste Alayrac, Radu Soricut, Angeliki Lazaridou, Orhan Firat, Julian Schrittwieser, et~al.
\newblock Gemini 1.5: Unlocking multimodal understanding across millions of tokens of context.
\newblock \emph{arXiv:2403.05530}, 2024.

\bibitem[Roziere et~al.(2023)Roziere, Gehring, Gloeckle, Sootla, Gat, Tan, Adi, Liu, Remez, Rapin, et~al.]{roziere2023code}
Baptiste Roziere, Jonas Gehring, Fabian Gloeckle, Sten Sootla, Itai Gat, Xiaoqing~Ellen Tan, Yossi Adi, Jingyu Liu, Tal Remez, J{\'e}r{\'e}my Rapin, et~al.
\newblock Code llama: Open foundation models for code.
\newblock \emph{arXiv:2308.12950}, 2023.

\bibitem[Shao et~al.(2024)Shao, Wang, Zhu, Xu, Song, Zhang, Li, Wu, and Guo]{shao2024deepseekmath}
Zhihong Shao, Peiyi Wang, Qihao Zhu, Runxin Xu, Junxiao Song, Mingchuan Zhang, YK~Li, Y~Wu, and Daya Guo.
\newblock Deepseekmath: Pushing the limits of mathematical reasoning in open language models.
\newblock \emph{arXiv:2402.03300}, 2024.

\bibitem[Tang et~al.(2024)Tang, Zhang, Wan, and Wei]{tang2024mathscale}
Zhengyang Tang, Xingxing Zhang, Benyou Wan, and Furu Wei.
\newblock Mathscale: Scaling instruction tuning for mathematical reasoning.
\newblock \emph{arXiv:2403.02884}, 2024.

\bibitem[Tong et~al.(2024)Tong, Li, Wang, Wang, Teng, and Shang]{tong2024can}
Yongqi Tong, Dawei Li, Sizhe Wang, Yujia Wang, Fei Teng, and Jingbo Shang.
\newblock Can llms learn from previous mistakes? investigating llms' errors to boost for reasoning.
\newblock \emph{arXiv:2403.20046}, 2024.

\bibitem[Toshniwal et~al.(2024)Toshniwal, Moshkov, Narenthiran, Gitman, Jia, and Gitman]{toshniwal2024openmathinstruct}
Shubham Toshniwal, Ivan Moshkov, Sean Narenthiran, Daria Gitman, Fei Jia, and Igor Gitman.
\newblock Openmathinstruct-1: A 1.8 million math instruction tuning dataset.
\newblock \emph{arXiv:2402.10176}, 2024.

\bibitem[Touvron et~al.(2023)Touvron, Lavril, Izacard, Martinet, Lachaux, Lacroix, Rozi{\`e}re, Goyal, Hambro, Azhar, et~al.]{touvron2023llama}
Hugo Touvron, Thibaut Lavril, Gautier Izacard, Xavier Martinet, Marie-Anne Lachaux, Timoth{\'e}e Lacroix, Baptiste Rozi{\`e}re, Naman Goyal, Eric Hambro, Faisal Azhar, et~al.
\newblock Llama: Open and efficient foundation language models.
\newblock \emph{arXiv:2302.13971}, 2023.

\bibitem[Tunstall et~al.(2023)Tunstall, Beeching, Lambert, Rajani, Rasul, Belkada, Huang, von Werra, Fourrier, Habib, et~al.]{tunstall2023zephyr}
Lewis Tunstall, Edward Beeching, Nathan Lambert, Nazneen Rajani, Kashif Rasul, Younes Belkada, Shengyi Huang, Leandro von Werra, Cl{\'e}mentine Fourrier, Nathan Habib, et~al.
\newblock Zephyr: Direct distillation of lm alignment.
\newblock \emph{arXiv:2310.16944}, 2023.

\bibitem[Wang et~al.(2023{\natexlab{a}})Wang, Ren, Zhou, Lu, Luo, Shi, Zhang, Song, Zhan, and Li]{wang2023mathcoder}
Ke~Wang, Houxing Ren, Aojun Zhou, Zimu Lu, Sichun Luo, Weikang Shi, Renrui Zhang, Linqi Song, Mingjie Zhan, and Hongsheng Li.
\newblock Mathcoder: Seamless code integration in llms for enhanced mathematical reasoning.
\newblock \emph{arXiv:2310.03731}, 2023{\natexlab{a}}.

\bibitem[Wang et~al.(2023{\natexlab{b}})Wang, Li, Shao, Xu, Dai, Li, Chen, Wu, and Sui]{wang2023math}
Peiyi Wang, Lei Li, Zhihong Shao, RX~Xu, Damai Dai, Yifei Li, Deli Chen, Y~Wu, and Zhifang Sui.
\newblock Math-shepherd: Verify and reinforce llms step-by-step without human annotations.
\newblock \emph{CoRR, abs/2312.08935}, 2023{\natexlab{b}}.

\bibitem[Wang et~al.(2023{\natexlab{c}})Wang, Xia, and Liu]{wang2023generative}
Zengzhi Wang, Rui Xia, and Pengfei Liu.
\newblock Generative ai for math: Part i--mathpile: A billion-token-scale pretraining corpus for math.
\newblock \emph{arXiv:2312.17120}, 2023{\natexlab{c}}.

\bibitem[Wei et~al.(2022)Wei, Wang, Schuurmans, Bosma, Xia, Chi, Le, Zhou, et~al.]{wei2022chain}
Jason Wei, Xuezhi Wang, Dale Schuurmans, Maarten Bosma, Fei Xia, Ed~Chi, Quoc~V Le, Denny Zhou, et~al.
\newblock Chain-of-thought prompting elicits reasoning in large language models.
\newblock \emph{NeurIPS}, 2022.

\bibitem[Xin et~al.(2024)Xin, Guo, Shao, Ren, Zhu, Liu, Ruan, Li, and Liang]{xin2024deepseek}
Huajian Xin, Daya Guo, Zhihong Shao, Zhizhou Ren, Qihao Zhu, Bo~Liu, Chong Ruan, Wenda Li, and Xiaodan Liang.
\newblock Deepseek-prover: Advancing theorem proving in llms through large-scale synthetic data.
\newblock \emph{arXiv:2405.14333}, 2024.

\bibitem[Xu et~al.(2024)Xu, Liu, Liu, Hou, Li, Zhang, Wang, Zeng, Du, Zhao, et~al.]{xu2024chatglm}
Yifan Xu, Xiao Liu, Xinghan Liu, Zhenyu Hou, Yueyan Li, Xiaohan Zhang, Zihan Wang, Aohan Zeng, Zhengxiao Du, Wenyi Zhao, et~al.
\newblock Chatglm-math: Improving math problem-solving in large language models with a self-critique pipeline.
\newblock \emph{arXiv:2404.02893}, 2024.

\bibitem[Yao et~al.(2024)Yao, Yu, Zhao, Shafran, Griffiths, Cao, and Narasimhan]{yao2024tree}
Shunyu Yao, Dian Yu, Jeffrey Zhao, Izhak Shafran, Tom Griffiths, Yuan Cao, and Karthik Narasimhan.
\newblock Tree of thoughts: Deliberate problem solving with large language models.
\newblock \emph{NeurIPS}, 2024.

\bibitem[Ying et~al.(2024)Ying, Zhang, Li, Zhou, Shao, Fei, Ma, Hong, Liu, Wang, et~al.]{ying2024internlm}
Huaiyuan Ying, Shuo Zhang, Linyang Li, Zhejian Zhou, Yunfan Shao, Zhaoye Fei, Yichuan Ma, Jiawei Hong, Kuikun Liu, Ziyi Wang, et~al.
\newblock Internlm-math: Open math large language models toward verifiable reasoning.
\newblock \emph{arXiv:2402.06332}, 2024.

\bibitem[Yoran et~al.(2023)Yoran, Wolfson, Bogin, Katz, Deutch, and Berant]{yoran2023answering}
Ori Yoran, Tomer Wolfson, Ben Bogin, Uri Katz, Daniel Deutch, and Jonathan Berant.
\newblock Answering questions by meta-reasoning over multiple chains of thought.
\newblock \emph{arXiv:2304.13007}, 2023.

\bibitem[Yu et~al.(2023)Yu, Jiang, Shi, Yu, Liu, Zhang, Kwok, Li, Weller, and Liu]{yu2023metamath}
Longhui Yu, Weisen Jiang, Han Shi, Jincheng Yu, Zhengying Liu, Yu~Zhang, James~T Kwok, Zhenguo Li, Adrian Weller, and Weiyang Liu.
\newblock Metamath: Bootstrap your own mathematical questions for large language models.
\newblock \emph{arXiv:2309.12284}, 2023.

\bibitem[Yuan et~al.(2023)Yuan, Yuan, Li, Dong, Tan, and Zhou]{yuan2023scaling}
Zheng Yuan, Hongyi Yuan, Chengpeng Li, Guanting Dong, Chuanqi Tan, and Chang Zhou.
\newblock Scaling relationship on learning mathematical reasoning with large language models.
\newblock \emph{arXiv:2308.01825}, 2023.

\bibitem[Yue et~al.(2023)Yue, Qu, Zhang, Fu, Huang, Sun, Su, and Chen]{yue2023mammoth}
Xiang Yue, Xingwei Qu, Ge~Zhang, Yao Fu, Wenhao Huang, Huan Sun, Yu~Su, and Wenhu Chen.
\newblock Mammoth: Building math generalist models through hybrid instruction tuning.
\newblock \emph{arXiv:2309.05653}, 2023.

\bibitem[Yue et~al.(2024)Yue, Zheng, Zhang, and Chen]{yue2024mammoth2}
Xiang Yue, Tuney Zheng, Ge~Zhang, and Wenhu Chen.
\newblock Mammoth2: Scaling instructions from the web.
\newblock \emph{arXiv:2405.03548}, 2024.

\bibitem[Zheng et~al.(2024)Zheng, Chiang, Sheng, Zhuang, Wu, Zhuang, Lin, Li, Li, Xing, et~al.]{zheng2024judging}
Lianmin Zheng, Wei-Lin Chiang, Ying Sheng, Siyuan Zhuang, Zhanghao Wu, Yonghao Zhuang, Zi~Lin, Zhuohan Li, Dacheng Li, Eric Xing, et~al.
\newblock Judging llm-as-a-judge with mt-bench and chatbot arena.
\newblock \emph{NeurIPS}, 2024.

\bibitem[Zhou et~al.(2022)Zhou, Sch{\"a}rli, Hou, Wei, Scales, Wang, Schuurmans, Cui, Bousquet, Le, et~al.]{zhou2022least}
Denny Zhou, Nathanael Sch{\"a}rli, Le~Hou, Jason Wei, Nathan Scales, Xuezhi Wang, Dale Schuurmans, Claire Cui, Olivier Bousquet, Quoc Le, et~al.
\newblock Least-to-most prompting enables complex reasoning in large language models.
\newblock \emph{arXiv:2205.10625}, 2022.

\bibitem[Zhou et~al.(2024)Zhou, Zhang, Wang, Chen, Zhao, Sha, Sheng, Wang, and Wen]{zhou2024jiuzhang3}
Kun Zhou, Beichen Zhang, Jiapeng Wang, Zhipeng Chen, Wayne~Xin Zhao, Jing Sha, Zhichao Sheng, Shijin Wang, and Ji-Rong Wen.
\newblock Jiuzhang3. 0: Efficiently improving mathematical reasoning by training small data synthesis models.
\newblock \emph{arXiv:2405.14365}, 2024.

\bibitem[Zhu et~al.(2024)Zhu, Guo, Shao, Yang, Wang, Xu, Wu, Li, Gao, Ma, et~al.]{zhu2024deepseek}
Qihao Zhu, Daya Guo, Zhihong Shao, Dejian Yang, Peiyi Wang, Runxin Xu, Y~Wu, Yukun Li, Huazuo Gao, Shirong Ma, et~al.
\newblock Deepseek-coder-v2: Breaking the barrier of closed-source models in code intelligence.
\newblock \emph{arXiv:2406.11931}, 2024.

\end{thebibliography}
\bibliographystyle{iclr2024_conference}


\end{document}